\documentclass[
]{ceurart}

\usepackage{listings}

\usepackage{subcaption}
\usepackage{pgf}
\usepackage{enumitem}
\usepackage{graphicx}
\usepackage{placeins}
\usepackage{wrapfig}
\usepackage[export]{adjustbox}

\usepackage{draftwatermark}
\SetWatermarkText{De-Factify 4.0 Workshop at AAAI 2025}
\SetWatermarkAngle{0}
\SetWatermarkScale{0.11}
\SetWatermarkVerCenter{0.95\paperheight}
\begin{document}

\copyrightyear{2025}
\copyrightclause{Copyright for this paper by its authors.
  Use permitted under Creative Commons License Attribution 4.0
  International (CC BY 4.0).}

\conference{De-Factify'25: 4\textsuperscript{th} Workshop on Multimodal Fact Checking and Hate Speech Detection,
Feb 25 -- Mar 4, 2025, Philadelphia, PA}

\title{SKDU at De-Factify 4.0: Vision Transformer with Data Augmentation for AI-Generated Image Detection}


\author[1]{Shrikant Malviya}[%
orcid=0000-0002-7539-3721,
email=s.kant.malviya@gmail.com,
url=https://skmalviya.bitbucket.io/,
]
\cormark[1]
\address[1]{Department of Computer Science, Durham University, UK}

\author[2]{Neelanjan Bhowmik}[%
orcid=0000-0002-7607-2209,
email=nilus.mail@gmail.com,
url=https://www.linkedin.com/in/neelanjanbhowmik/,
]

\address[2]{British Car Auctions, UK}

\author[1]{Stamos Katsigiannis}[%
orcid=0000-0001-9190-0941,
email=stamos.katsigiannis@durham.ac.uk,
url=https://www.durham.ac.uk/staff/stamos-katsigiannis/,
]

\cortext[1]{Corresponding author.}

\vspace{-0.2cm}
\begin{abstract}
The aim of this work is to explore the potential of pre-trained vision-language models, e.g. Vision Transformers (ViT), enhanced with advanced data augmentation strategies for the detection of AI-generated images. Our approach leverages a fine-tuned ViT model trained on the Defactify-4.0 dataset, which includes images generated by state-of-the-art models such as Stable Diffusion 2.1, Stable Diffusion XL, Stable Diffusion 3, DALL-E 3, and MidJourney. We employ perturbation techniques like flipping, rotation, Gaussian noise injection, and JPEG compression during training to improve model robustness and generalisation. The experimental results demonstrate that our ViT-based pipeline achieves state-of-the-art performance, significantly outperforming competing methods on both validation and test datasets.
\end{abstract}


\maketitle

\vspace{-1cm}
\section{Introduction} \label{intro}
\begin{wrapfigure}{r}{0.4\linewidth}
    \centering
    \includegraphics[width=\linewidth]{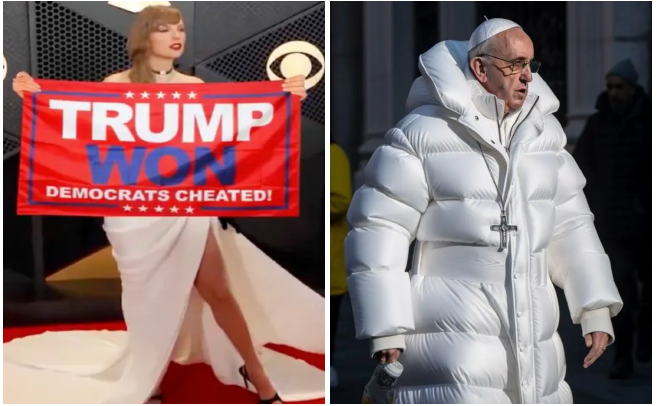}
    \vspace{-0.8cm}
    \caption{Example fake images generated by AI generative models.}
    \label{img:fake_examples}
\end{wrapfigure}

\vspace{-0.24cm}
Recent advancements in generative models have revolutionised the creation of photo-realistic images, making it faster, cheaper, and more accessible than ever before. Techniques such as Generative Adversarial Networks (GANs) and diffusion models have enabled unprecedented realism in synthetic images, reshaping industries from entertainment to advertising \cite{cardenuto2023age,barrett2023identifying}. Moreover, with readily accessible, pre-trained text-to-image models, like DALL-E, MidJourney, and Stable Diffusion, have made it easier to generate high-quality synthetic images. Examples (Figure~\ref{img:fake_examples}), such as ``The Pope in Drip'' \cite{edwards2023immaculate}, ``Taylor Swift `Endorsing' Donald Trump'' \cite{novakviral}, demonstrate how these synthetic creations increasingly blend with natural images, leading to two key types of misidentifications: synthetic images being mistaken for natural ones and natural images being misclassified as synthetic.

Synthetic images are increasingly difficult to distinguish from natural images, resulting in misidentification cases that can have far-reaching consequences. Harmless examples, such as synthetic artwork falsely attributed to real artists highlight ethical dilemmas around authorship and intellectual property \cite{nightingale2022aisynthesized}. More alarmingly, synthetic media have been implicated in disinformation campaigns, deepfakes, and targeted misinformation, where fabricated visuals lend credibility to false narratives. A notable example includes the $2018$ Gabon political crisis, where suspicions over the authenticity of a video exacerbated political unrest \cite{2019welcome}. 
These incidents underline the urgent need for robust detection systems to reliably distinguish real from synthetic content \cite{cooke2024good}.



AI-generated image detection (AGID) has become a critical area of research due to the rapid advancements in generative models such as GAN and diffusion models. Early forensic approaches relied on identifying peculiar traces, termed artificial fingerprints, left by synthesis models in generated images \cite{wissmann2024whodunit}. These subtle artifacts, first observed in GAN-based methods, provide essential clues for detection tasks. However, the shift towards diffusion models has introduced new challenges, as their generation artifacts differ significantly from those of GANs \cite{corvi2023intriguing,li2024improving}. Traditional CNN-based classifiers, such as those based on ResNet-50 \cite{he2016deep}, have shown promise in generalising to out-of-distribution (OOD) data when trained on large and diverse datasets with extensive augmentation strategies. For example, training on a dataset of 360k ProGAN images enabled significant generalisation to other GAN-based generators, as noted in \cite{wang2020cnngenerated,sinitsa2024deep}. Despite this, their performance on diffusion-based methods remains limited, underscoring the need for more adaptable and robust solutions.

To address these issues, this paper explores advanced strategies for improving AGID generalisation performance. Recently, Vision Transformers (ViTs) have emerged as a powerful architecture for image classification and detection tasks \cite{ojha2023universal,cozzolino2024raising}, offering distinct advantages over convolutional networks. Unlike ResNet-based architectures, ViTs utilise a self-attention mechanism, which allows them to capture long-range dependencies and global features effectively. This capability is particularly beneficial for in-distribution (ID) scenarios, where fine-grained visual features are crucial for distinguishing real images from AI-generated ones. Pre-trained ViT models, fine-tuned on domain-specific datasets, have demonstrated state-of-the-art performance in detecting subtle generative artifacts within ID data \cite{moskowitz2024detecting,liu2024forgeryaware}. Moreover, the adaptability of ViTs, combined with advanced data augmentation techniques such as perturbations and normalisation, enhances their robustness to post-processing transformations such as compression and resizing, which often obscure detection-critical traces \cite{li2024improving}.

We emphasise the need for feature-sensitive training paradigms that preserve subtle artifacts and employ diverse augmentation techniques to bridge distributional disparities between training and testing data. By investigating the interplay between universal artifact features and training methodologies, our work contributes to the development of robust AGID systems capable of adapting to the rapidly evolving landscape of generative models. We make our code publicly available at \url{ https://github.com/skmalviya/ai_gen\_image\_defactify}. In summary, the key contributions of this work are as follows: 

\begin{itemize}[noitemsep,topsep=0pt,leftmargin=*]
    \item[--] We showcase the efficacy of pre-trained ViTs, which utilise self-attention mechanisms for superior feature extraction, highlighting their edge over conventional CNN architectures in distinguishing fine-grained generative artifacts.
    \item[--] To show the effectiveness of data augmentation techniques, our pipeline incorporates perturbations such as flipping, brightness adjustments, Gaussian noise, and JPEG compression to simulate real-world distortions, ensuring robustness to diverse challenges in AI-generated image detection.
    \item[--] We present an extensive evaluation of our approach on the Defactify-4.0 dataset, achieving superior results compared to traditional methods and robustness under the in-distribution scenario.
    
\end{itemize}

\vspace{-0.4cm}
\section{Shared Task CT\textsuperscript{2}: AI-Generated Image Detection} \label{method}
\vspace{-0.2cm}
The \textit{CT\textsuperscript{2}: AI-Generated Image Detection} challenge introduces a comprehensive dataset containing images, associated captions, and detailed labels to facilitate classification tasks. 
\vspace{-0.4cm}
\subsection{Tasks}
\vspace{-0.1cm}
The dataset is designed to support two key classification tasks:

\begin{itemize}[noitemsep,topsep=0pt,leftmargin=*]
    \item[--] \textbf{Task A}: Classifying images as either real or AI-generated. This binary classification task evaluates a model's ability to differentiate between authentic and synthetic content, a critical capability in combating misinformation and ensuring media authenticity.
    \item[--] \textbf{Task B}: Identifying the specific generative AI model responsible for generating synthetic images. The generative models are: \{\textit{Stable Diffusion 3 (SD-3), Stable Diffusion XL (SD-XL), Stable Diffusion 2.1 (SD-2.1), DALL-E 3, Midjourney 6}\}. This multi-class classification task focuses on determining the generative source, which is essential for understanding the characteristics and behaviours of different models and for developing tailored detection strategies.
\end{itemize}

\begin{figure}[!htb]
    \centering
    \includegraphics[width=\linewidth]{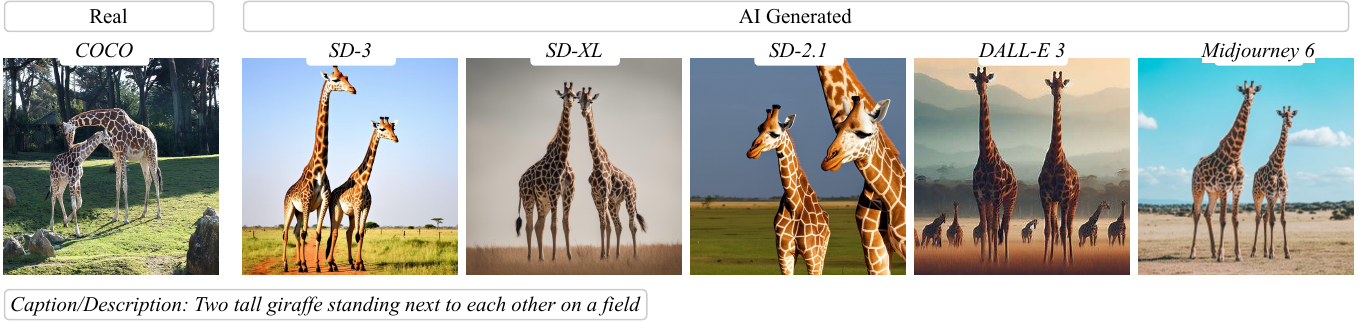}
    \vspace{-0.4cm}
    \caption{Exemplar images from Defactify-4.0's: \{\textit{Real vs AI generated}\} classification (\textbf{Task A}), detect the AI generated models (\textbf{Task B}).}
    \vspace{-0.8cm}
    \label{img:de_db}
\end{figure}

\vspace{-0.8cm}
\subsection{Dataset}
\vspace{-0.2cm}
\label{sec:data_desc}
The Defactify-4.0's image dataset includes images generated by different generative models as described above. The dataset\footnote{Dataset link: https://huggingface.co/datasets/NasrinImp/Defactify4\_Train} consists of $50$K samples, developed based on the MS-COCO \cite{lin2014microsoft} dataset. The process involves taking the original (real) MS-COCO captions and images and feeding the captions into each of the different generative models. Hence, the dataset consists of six classes of images: COCO images (real), and five AI-generated classes (Figure~\ref{img:de_db}). To elevate the competition further, a new bigger testing data comprising modified images with various perturbations, e.g. Horizontal flipping, Brightness Reduction, Gaussian Noise, and JPEG Compression, is shared for evaluating both tasks.

\vspace{-0.4cm}
\section{Methodology}
\vspace{-0.2cm}
\subsection{Our Pipeline}
\begin{figure}[!htb]
  \centering
  \includegraphics[width=1\textwidth]{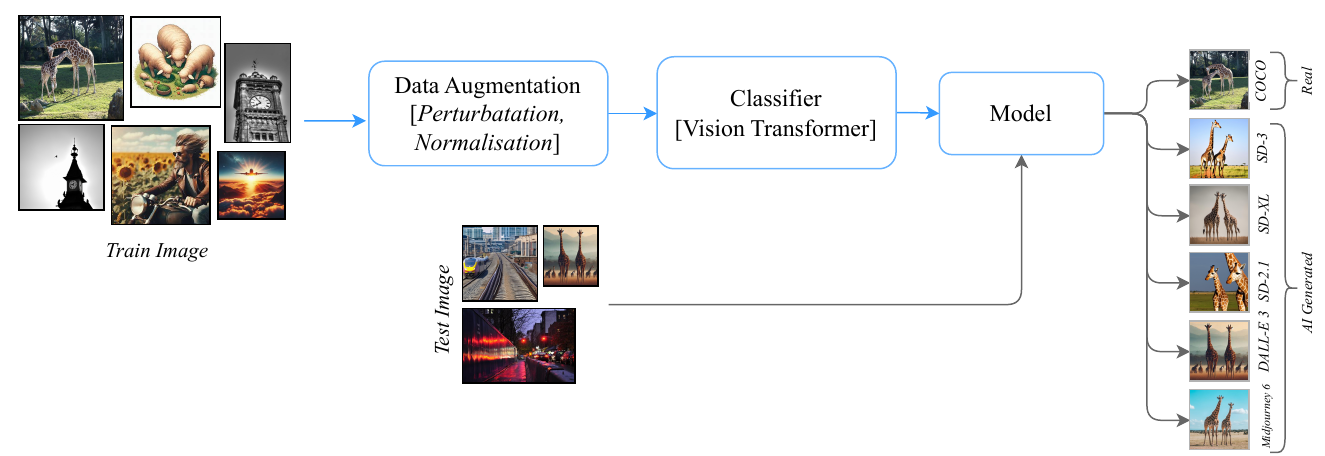}
  \vspace{-0.6cm}
  \caption{Overview of the Training/Testing pipeline to process images for the classification tasks.}
  \vspace{-0.6cm}
  \label{img:proposed_pipeline}
\end{figure}

\vspace{-0.2cm}
The pipeline illustrated in Figure~\ref{img:proposed_pipeline} outlines a systematic approach for training and testing a classification model designed to differentiate between real images and those generated by AI models. The framework comprises two stages, training and testing, each utilising distinct image preprocessing steps. During the training, the pipeline uses a pre-trained Vision Transformer (ViT) \cite{dosovitskiy2021image} architecture on the provided training dataset, while testing uses the trained model to classify into six classes (Section~\ref{sec:data_desc}).

\vspace{-0.4cm}
\subsection{Training stage}
\vspace{-0.2cm}
In the training phase, input images are first processed through an image preprocessing pipeline that incorporates \textit{perturbation} and \textit{normalisation} techniques. Perturbation refers to data augmentation methods such as flipping, cropping, rotation, and adding noise to enhance the model's robustness and generalisation ability. Normalisation ensures that pixel values are scaled uniformly, improving convergence during model training.

\begin{figure}[!htb]
    \centering
    \includegraphics[width=\linewidth]{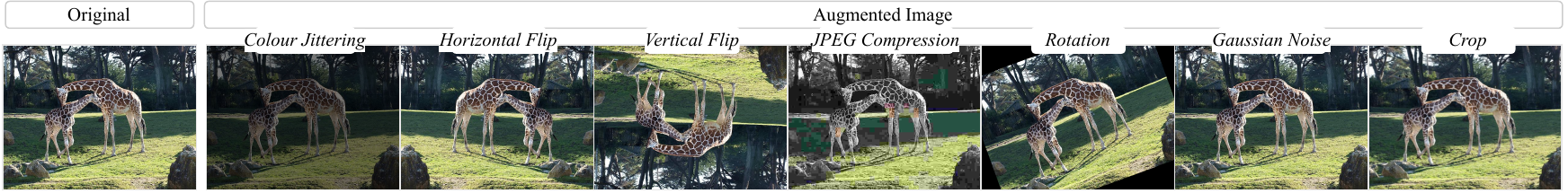}
    \vspace{-0.4cm}
    \caption{Exemplar images from Defactify-4.0's: During model training, various augmentation techniques are applied to the input images.}
    \vspace{-0.6cm}
    \label{img:aug_ex}
\end{figure}

\vspace{-0.4cm}
\subsubsection{Model}
\vspace{-0.2cm}
The preprocessed images are then passed through a \textit{pre-trained Vision Transformer (ViT)} model. The ViT divides each image into a sequence of fixed-size patches, which are subsequently embedded into a lower-dimensional space and treated as input tokens for the transformer architecture. The attention mechanism of the model enables it to capture both local and global dependencies within the image, facilitating the extraction of meaningful features relevant to the task. This pre-trained model serves as the foundation for learning, leveraging its prior knowledge of general image features from large-scale datasets. During this phase, the model is fine-tuned to adapt its feature extraction capabilities to the specific task of classifying images into one of six classes.

\vspace{-0.4cm}
\subsubsection{Data Augmentation} \label{sec:data_aug}
\vspace{-0.2cm}
With the release of the new challenging testing data, the simple ViT-based classifier struggles with distorted images (e.g., Brightness Reduction, Gaussian Noise, JPEG Compression, etc.). This indicates that the model lacks robustness to these transformations. To make the model robust to distortions, we opted to include several transformations in the training pipeline.  By training on augmented data, the model learns to generalise to distorted inputs. The following augmentations are employed:

\begin{itemize}[noitemsep,topsep=0pt,leftmargin=*]
\item[--] \textbf{Colour Jittering:} Random brightness adjustments in the range of $0.45-0.55$ introduce variability in pixel intensities.
\item[--] \textbf{Random Horizontal/Vertical Flips:} Probabilistic mirroring of images to increase sample diversity.
\item[--] \textbf{JPEG Compression:} Images are subjected to compression with a quality range of $30-70$ to simulate real-world compression noise.
\item[--] \textbf{Random Rotation:} Images are randomly rotated within the range of $0-90$ degrees to introduce perspective variations.
\item[--] \textbf{Gaussian Noise:} Additive noise with mean $0$ and standard deviation up to $0.3$ is applied to simulate sensor noise.
\item[--] \textbf{Random Crop:} Images are cropped and resized within a predefined range to simulate different zoom levels and perspectives.
\end{itemize}



\vspace{-0.4cm}
\subsection{Testing stage}
\vspace{-0.2cm}
During testing, the input images undergo a simplified preprocessing pipeline that involves only \textit{normalisation}, ensuring consistency in input scaling. The normalised images are then fed into the \textit{fine-tuned ViT model}, which has been optimised for the specific classification task. The fine-tuned ViT predicts the class of each image based on the learned features, distinguishing between real images and those generated by various AI models. The testing phase evaluates the model's ability to generalise to unseen data without the influence of data augmentation or perturbation.
\vspace{-0.4cm}
\section{Results \& Discussion} \label{results}
\vspace{-0.2cm}
\subsection{Evaluation Experiments}
\vspace{-0.2cm}

We evaluated three versions of our ViT-based model on Defactify-4.0's validation and test sets using F1-scores for Tasks A and B: (a) Plain ViT (Section~\ref{method}), (b) CNN $\rightarrow$ ViT Classifier, and (c) CNN $\oplus$ ViT Classifier. For comparison, we also tested the SAFE model \cite{li2024improving}, a CLIP-based classifier \cite{moskowitz2024detecting}, Whodunit~\cite{wissmann2024whodunit}, and CNN-based ResNet-50 \cite{he2016deep}.

\vspace{-0.2cm}
\subsection{Implementation Details}
\vspace{-0.1cm}
We fine-tune the pre-trained ViT model (\texttt{google/vit-base-patch16-224}\footnote{\url{https://huggingface.co/google/vit-base-patch16-224}}) on the task dataset using proposed augmentations during training (with application probabilities of $0.2 - 0.5$), and resizing/normalisation for testing. The model is trained with AdamW (learning rate (lr): $2e\!-\!5$, weight decay: $0.01$), a batch size of $128$, and $15$ epochs, with Accuracy/F1-score as the evaluation metric. Training time per epoch was 40 minutes, leads to total training time (15 epochs): 10 hours approx.


To combine convolution features with ViT, we use two strategies: 1) CNN $\rightarrow$ ViT Classifier, where CNN features are upscaled and encoded by ViT, and 2) CNN $\oplus$ ViT Classifier, where CNN and ViT features are concatenated before classification. All models use the same preprocessing module with proposed augmentations (Section~\ref{sec:data_aug}).


We compare methods using ResNet-50 \cite{he2016deep}, pre-trained on ImageNet and fine-tuned with the same strategy as our ViT-based model. The SAFE model \cite{li2024improving} employs a lightweight ResNet (1.44M parameters) with specific augmentations and trains using AdamW (batch size $32$, lr $5e\!-\!3$). The CLIP-based classifier (\texttt{openai/clip-vit-large-patch14}\footnote{\url{https://huggingface.co/openai/clip-vit-large-patch14}}) uses a ViT-L/14 encoder with a sequential classification head, trained with AdamW (batch size $16$, lr $1e\!-\!6$). Whodunit \cite{wissmann2024whodunit} employs an AlexNet-like architecture with PSD features and trains over $40$ epochs (batch size $16$, lr $1e\!-\!4$, $75\%$ dropout). 
All experiments were conducted on NVIDIA GeForce RTX 4090 type GPUs.

\begin{table*}
\renewcommand{\arraystretch}{0.85}
\centering
  \caption{Performance (F1-score) on the development set and test set.}
  \label{tab:main_result}
  \begin{tabular}{lcccc}
    \toprule

\multirow{2}{*}{Models} & \multicolumn{2}{c}{ Validation set } & \multicolumn{2}{c}{ Test set} \\
\cmidrule(lr){2-3} 
\cmidrule(lr){4-5}
    
     & F1 (Task-A) & F1 (Task-B) & F1 (Task-A) & F1 (Task-B)\\
    \midrule
    CLIP Classifier \cite{moskowitz2024detecting} & 0.5268 & 0.1667 & 0.5148 & 0.1547 \\
    SAFE \cite{li2024improving} & 0.5452 & 0.1913 & 0.5246 & 0.1762\\
    Whodunit (PSD) \cite{wissmann2024whodunit} & 0.7188 & 0.6533 & 0.7340 & 0.1978\\
    CNN (ResNet-50) & 0.9636 & 0.9227 & 0.7987 & 0.3928 \\
    \hline
    CNN $\rightarrow$ ViT Classifier & 0.9868 & 0.9664 & 0.8158 & 0.4299 \\
    CNN $\oplus$  ViT Classifier & 0.9827 & 0.9778 & 0.8200 & 0.4310 \\
    ViT Classifier & \textbf{0.9963} & \textbf{0.9839} & \textbf{0.8292} & \textbf{0.4864}\\
  \bottomrule
  \vspace{-0.6cm}
\end{tabular}
\end{table*}

\vspace{-0.4cm}
\subsection{Results}
\vspace{-0.1cm}
Our ViT-based model achieves state-of-the-art performance, as highlighted by its F1 scores on both validation and test sets (Table~\ref{tab:main_result}). Compared to baseline methods like CLIP, SAFE, and Whodunit, our model demonstrates superior accuracy. For F1-score (Defactify4.0 challenge score), we achieved a validation score of 0.9963 and 0.9839 for Task-A and Task-B respectively on the augmented validation set. On the other hand, we obtained a score of 0.8292 and 0.4864 on the testing leaderboard. These results underscore the effectiveness of ViT in extracting fine-grained features essential for distinguishing real and AI-generated content. 
To show the effectiveness of our data augmentation, we also train ViT without any perturbation resulting low F1-score of 0.7190 and 0.1856, respectively.
We also observe that using the standalone CNN-based (ResNet-50) classifier or incorporating its CNN features into the ViT pipeline, either by concatenating CNN and ViT outputs or sequentially feeding CNN features to ViT, do not yield any substantial improvement in performance compared to using ViT alone. 

The significant performance gap between CLIP and ViT in the classification task stems primarily from their pre-training objectives. CLIP is designed for a contrastive learning objective, aligning images and text embeddings in a shared space, which makes it suitable for tasks like zero-shot classification or image-text retrieval. However, this generalisation can make CLIP less effective for fine-grained classification tasks. In contrast, ViT models are pre-trained specifically for image classification tasks on datasets such as ImageNet, making their embeddings more tailored for this purpose. This gives ViT a natural advantage in tasks requiring nuanced visual discrimination, such as distinguishing between AI-generated and real images.

While our ViT-based approach shows strong generalization within the given dataset, we acknowledge that its performance on unseen generative models remains an open question. In future, We will discuss how self-attention in ViTs enables better generalisation by capturing high-level patterns rather than local textures. We would also like to explore on zero-shot learning using pre-trained CLIP models or hybrid architectures to improve generalisation.
\vspace{-0.3cm}
\section{Conclusion} 
\vspace{-0.2cm}
This work introduces a robust ViT-based pipeline for detecting AI-generated images, leveraging advanced augmentations and pre-trained ViT models for state-of-the-art performance across diverse generative methods. Perturbation strategies enhance resilience to distortions like JPEG compression and noise, ensuring superior generalisation and robustness, outperforming existing methods on in-distribution detection. The pipeline's reliance on large pre-trained models poses challenges in computational overhead and resource needs. Its effectiveness on out-of-distribution synthetic images remains to be tested. Future work will focus on lightweight models and adaptive learning to reduce resource dependency while ensuring performance, advancing robust solutions for AI-generated media.



\vspace{-0.3cm}
\section{Acknowledgments}
The authors in this project have been funded by UK EPSRC grant ``AGENCY: Assuring Citizen Agency in a World with Complex Online Harms'' under grant EP/W032481/2.

This work used the DiRAC@Durham facility managed by the Institute for Computational Cosmology on behalf of the STFC DiRAC HPC Facility (www.dirac.ac.uk). The equipment was funded by BEIS capital funding via STFC capital grants ST/P002293/1, ST/R002371/1 and ST/S002502/1, Durham University and STFC operations grant ST/R000832/1. DiRAC is part of the National e-Infrastructure.





\bibliography{sample-ceur}

\end{document}